\documentclass[letterpaper, 10 pt, conference]{ieeeconf}  

\IEEEoverridecommandlockouts                              

\usepackage{graphicx} 
\graphicspath{{media/}} 
\usepackage{subfigure}
\usepackage{algorithm}
\usepackage{algorithmic}

\usepackage{hyperref}
\usepackage{soul}
\usepackage{tabularx,booktabs}
\usepackage{leftidx}
\usepackage{cancel}

\usepackage{mathtools}
\usepackage{amsmath} 
\usepackage{amssymb}  
\usepackage{multirow}
\usepackage{cite}

\usepackage{color}
\newcolumntype{M}[1]{>{\centering\arraybackslash}m{#1}}

\title{\LARGE \bf Picking by Tilting: In-Hand Manipulation for\\Object Picking using Effector with Curved Form}

\author{Yanshu Song, Abdullah Nazir, Darwin Lau, and Yun-Hui Liu
\thanks{Y. Song, A. Nazir, D. Lau, and Y. Liu are with The Chinese University of Hong Kong, Hong Kong.}
}

\begin{document}

\maketitle
\thispagestyle{empty}
\pagestyle{empty}
\begin{abstract}
This paper presents a robotic in-hand manipulation technique that can be applied to pick an object too large to grasp in a prehensile manner, by taking advantage of its contact interactions with a curved, passive end-effector, and two flat support surfaces.
First, the object is tilted up while being held between the end-effector and the supports.
Then, the end-effector is tucked into the gap underneath the object, which is formed by tilting, in order to obtain a grasp against gravity.
In this paper, we first examine the mechanics of tilting to understand the different ways in which the object can be initially tilted.
We then present a strategy to tilt up the object in a secure manner.
Finally, we demonstrate successful picking of objects of
various
size and geometry using our technique through a set of experiments performed with a custom-made robotic device and a conventional robot arm.
Our experiment results show that object picking can be performed reliably with our method using
simple hardware and control, and when possible, with appropriate fixture design.

\end{abstract}

\section{Introduction}

\begin{figure}[t]
    \centering
    \includegraphics[trim={0cm 0cm 0cm 0cm}, clip, width=3.2in]{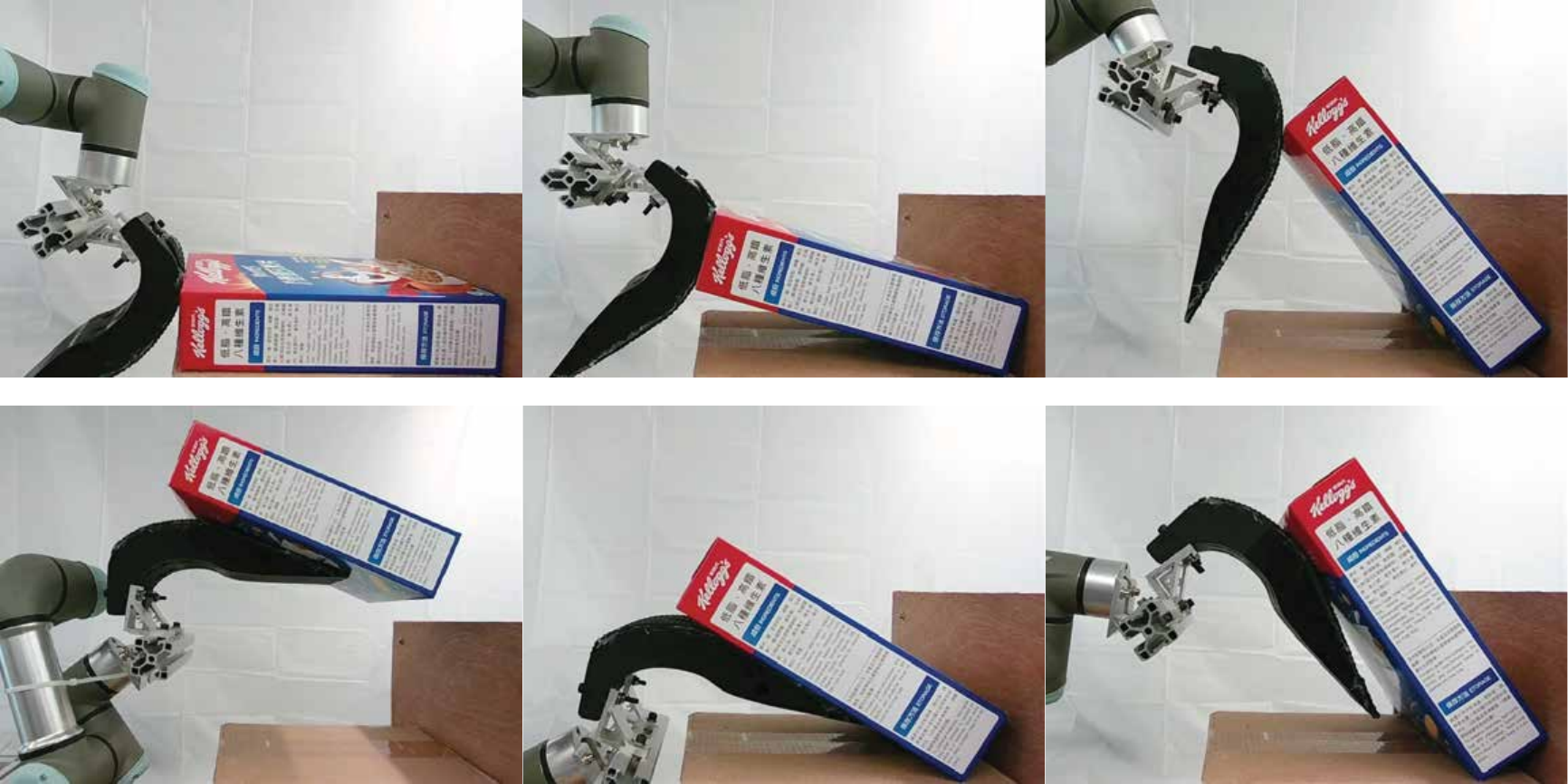}
    \caption{Sequence of snapshots (clockwise from the top-left) showing our picking technique performed with a curved, passive end-effector.}
    \label{fig:intro}
\end{figure}

This paper investigates a robotic in-hand manipulation technique for picking objects using a curved, passive end-effector. It is targeted at objects placed on a flat surface like a tabletop and physically blocked on one side by a wall-like support.
See Fig.~\ref{fig:intro} (clockwise from the top-left) showing the progress of the manipulation process.
First, the end-effector makes contact on a side of the object with its curved surface. Next, the object, which is in rolling contact with the end-effector, is tilted up while being pressed against the wall-like support.
Meanwhile, the end-effector is also reorienting itself so that at the end of tilting (third panel), it can block the object from escaping.
Subsequently, the end-effector gets into
the gap created by tilting in order to support the object from its bottom. Finally, the object is detached from the supporting surfaces and a grasp closed under gravity is obtained.

The proposed picking technique can be considered a practically useful material handling capability as it can enable robots to acquire grasps without prehension using low-cost, simple end-effectors.
This makes it applicable to objects that are too large to apply more straightforward methods such as grasping using a parallel-jaw gripper.
However, successful object picking with our method can also be considered more challenging as it is contingent on the ability to adequately manage the contact forces and the contact modes of interaction between the object, the end-effector and the supporting surfaces.

Following a formal problem description in Sec.~\ref{sec:problem_description} in which the task of object picking is formulated as a quasistatic process in a two-dimensional plane, in Sec.~\ref{sec:mechanics_planning} we analyze the mechanics and planning of tilting, which is a critical step in our picking technique. The mechanics of tilting concerns two distinct modes in which the object model can be initially tilted, while planning for tilting provides a blueprint to securely tilt the object by following a sequence of statically stable configurations.
In Sec.~\ref{sec:experiments}, the viability and practicality of our method is demonstrated through a set of experiments performed with a custom-made robotic device and a conventional manipulator arm.

\section{Related Work}

Our work is concerned with robotic in-hand manipulation, which means the capability to reconfigure an object relative to the robotic hand. See \cite{897777} for a general introduction to in-hand manipulation techniques.
One classical approach is to use high degrees-of-freedom (DOF) devices such as the three-fingered, nine-DOF Salisbury Hand \cite{mason1985robot}, featuring large-curvature fingertips that interact with the object through nonsliding/nonrolling contacts.
In contrast, in this work in-hand manipulation is achieved with zero-DOF, rigid body end-effector by exploiting environmental surfaces and sliding/rolling contacts with the object.
Previously, zero-DOF end-effectors have been used for object picking \cite{973345}, \cite{mucchiani2018object}, object transport \cite{specian2018robotic}, \cite{9689055} and stable object pose reconfiguration \cite{doi:10.1177/027836499801700502}, \cite{9196976}.
Fixed environmental surfaces have been exploited for grasping \cite{doi:10.1177/0278364914559753}, assembly \cite{Kim2019Shallow} and in-hand repositioning \cite{Dafle2014extrinsic}.
A growing body of studies, including some of the early works \cite{doi:10.1177/027836499000900302, 614264, doi:10.1177/027836402321261968}, have affirmed the importance of rolling/sliding contacts for manipulation dexterity. Recent studies to confirm this include \cite{9197146} which adopts active rolling surfaces in a gripper and \cite{7913727} which realizes in-hand sliding by exploiting inertial loads.

Considering that the presented
technique
does not firmly grasp the object, our work is also relevant to nonprehensile manipulation, which refers to ``manipulation without grasping'' \cite{Mason-1999-15061}. Notable examples of nonprehensile manipulation are pushing \cite{Lynch_1992},  pivoting \cite{583091} and toppling \cite{Lynch_1999}. Our picking technique features tilting, which can be considered inverse of toppling.
\cite{8794366} presents a motion-force control scheme for tilting, realized using a high fidelity force/torque sensor.
A tilting-based picking technique applicable to thin objects is presented in \cite{9197493}.



\section{Problem Description}
\label{sec:problem_description}
To pick an object for which a prehensile grasp may not be admissible,
Fig.~\ref{fig:intro} shows that it is workable to obtain a ``grasp against gravity'', as when a waiter supports a tray on a palm, by taking advantage of resources extrinsic to the object such as its contacts with the environmental surfaces.
In this work, our goal is to verify that the dexterous interactions necessary to accomplish this way of grasping can be realized using a simple, curved rigid body end-effector---henceforth referred to as the {\em palm}.

\begin{figure*}[t]
    \centering
    \subfigure[]{
    \includegraphics[trim={0cm 0cm 0cm 0cm}, clip, height=1.75in]{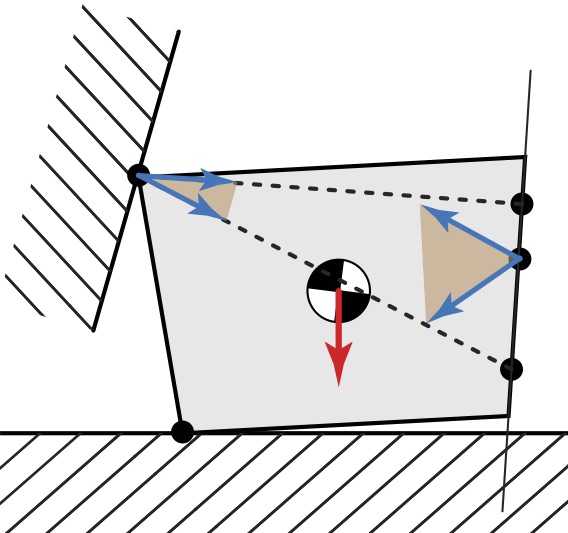}}
    \hspace{5mm}
    \subfigure[]{
    \includegraphics[trim={0cm 0cm 0cm 0cm}, clip, height=1.75in]{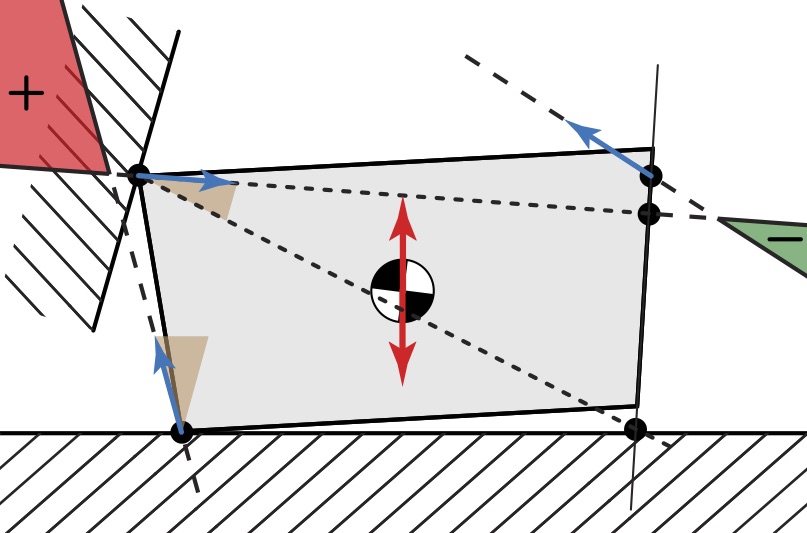}}
    \hspace{5mm}
    \subfigure[]{
    \includegraphics[trim={0cm 0cm 0cm 0cm}, clip, height=1.75in]{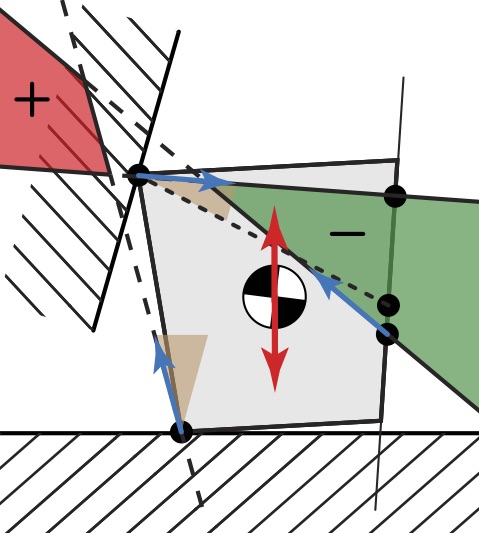}}
    \setlength{\unitlength}{1cm}
    \begin{picture}(0,0)(12,0.7)
    \put(-3.02,1.1){\footnotesize support \#1}
    \put(-4.62,4.95){\footnotesize support \#2}
    \put(-1.72,4.61){\footnotesize line of edge}
    \put(-4.23,1.60){\footnotesize $A$}
    \put(-4.21,3.76){\footnotesize $B$}
    \put(-0.96,2.87){\footnotesize $C$}
    \put(-0.96,3.36){\footnotesize $B'_1$}
    \put(-1.06,1.95){\footnotesize $B'_2$}
    \put(-2.51,2.01){\footnotesize $m\mathbf{g}$}
    \end{picture}
    \begin{picture}(0,0)(5.1,0.3)
    \put(-3.12,0.71){\footnotesize support \#1}
    \put(-6.03,4.55){\footnotesize support \#2}
    \put(-2.12,4.25){\footnotesize line of edge}
    \put(-5.66,1.20){\footnotesize $A$}
    \put(-5.62,3.33){\footnotesize $B$}
    \put(-1.32,3.28){\footnotesize $C$}
    \put(-1.41,2.67){\footnotesize $B'_1$}
    \put(-1.49,1.25){\footnotesize $B'_2$}
    \put(-3.39,1.65){\footnotesize $m\mathbf{g}$}
    \put(-4.24,2.89){\footnotesize $-m\mathbf{g}$}
    \put(-6.0,1.63){\footnotesize $\mathbf{f}^{A_1}$}
    \put(-5.25,3.33){\footnotesize $\mathbf{f}^{B_1}$}
    \put(-1.97,3.70){\footnotesize $\mathbf{f}_{C}$}
    \end{picture}
    \begin{picture}(0,0)(-2.6,0.7)
    \put(-5.38,1.11){\footnotesize support \#1}
    \put(-6.26,4.95){\footnotesize support \#2}
    \put(-4.47,4.55){\footnotesize line of edge}
    \put(-5.89,1.60){\footnotesize $A$}
    \put(-5.92,3.37){\footnotesize $B$}
    \put(-3.86,1.98){\footnotesize $C$}
    \put(-3.73,3.61){\footnotesize $B'_1$}
    \put(-3.76,2.51){\footnotesize $B'_2$}
    \put(-4.68,2.01){\footnotesize $m\mathbf{g}$}
    \put(-5.02,3.39){\footnotesize $-m\mathbf{g}$}
    \put(-6.24,2.03){\footnotesize $\mathbf{f}^{A_1}$}
    \put(-5.42,3.75){\footnotesize $\mathbf{f}^{B_1}$}
    \put(-4.33,2.37){\footnotesize $\mathbf{f}_{C}$}
    \end{picture}
    \caption{
    Feasibility of initial tilting with two contacts (a) and three contacts (b-c). In (a), the object is wedged between contacts $B$ and $C$. Object tilts by rotating about the stationary contact $B$ so contact $A$ can break free. When tilting with three contacts (b-c), the object slides to the right of the inward contact normal (not shown) at $A$ and $B$.
    This manner of tilting is feasible in (b) because the wrench of gravity can be quasistatically balanced by the composite wrench cone (represented using moment labels) of the three contact wrenches. It is not feasible in (c) because quasistatic balance of contact and gravity wrenches cannot be attained. 
    }
    \label{fig:mechanics}
\end{figure*}
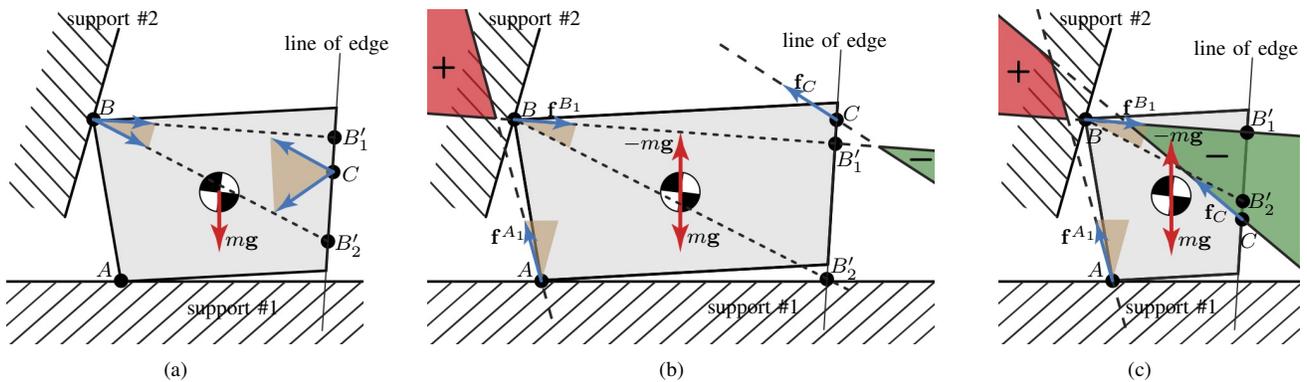

We assume that the object of interest is initially placed on a flat surface, denoted support \#1, which is perpendicular to the line of gravity.
Another flat surface, denoted support \#2 and not necessarily perpendicular to support \#1, blocks the object on one of its sides.  
The manipulation task is modeled as quasistatic process in a plane normal to the two supports.
As can be previewed in Fig.~\ref{fig:mechanics}, the target {\em object} is modeled as trapezoid whose top edge is possibly longer, but not shorter, than its bottom edge. It can be extruded or revolved into an inverted truncated cone, which represents the class of three-dimensional objects on which our method is applicable.
It is assumed that the planar object model makes frictional point contact with support \#1 at $A$, with support \#2 at $B$, and with the palm at $C$. It is further assumed that the object and its environment, comprising the palm and the support surfaces, interact according to the rules of rigid body mechanics. 

\section{Mechanics and Planning of Tilting}
\label{sec:mechanics_planning}

This section investigates the mechanics of tilting and presents a strategy to securely tilt the object.

\subsection{Mechanics}
\label{subsec:mechanics}

We examine two ways in which the object model can be
tilted
near its initial configuration
by forces applied to $C$ through the palm.
Let $B'_1$ ($B'_2$) be the image of $B$ on the
line of the edge that contains $C$,
under the left (right) edge of its friction cone as shown in Fig.~\ref{fig:mechanics}.
In the following paragraphs, the tilting behavior of the object model is characterized by
the order in which the points $B'_1$, $B'_2$ and $C$ occur on the line of edge when it is traversed from top to bottom.

\subsubsection{$\{B'_1CB'_2\}$}
In this configuration (Fig.~\ref{fig:mechanics}(a)), $C$ is in the interior of the line segment $\overline{B'_1B'_2}$. In other words, it is contained in $B$'s friction cone.
Now, if the friction cone at $C$ is large enough to contain $B$, then the two friction cones can ``see'' each other by their line of sight $\overline{BC}$. Accordingly, the object is in force-closure or ``wedged'' between the palm and support \#2, similarly to a peg stuck during peg-in-hole insertion task \cite{Mason__2001_3803}.
This configuration allows for a two-contact tilting: an upward palm motion will instantaneously rotate the object counterclockwise about the point coincident with $B$ and contact $A$ will break free from support \#1.

If $C$ is not in the interior of the line segment $\overline{B'_1B'_2}$,
force-closure with only two frictional contacts $B$ and $C$ cannot be attained because their line of sight $\overline{BC}$ will not be contained in $B$'s friction cone. Therefore, the two-contact tilting described above will not happen.
Instead, we will check the feasibility of a three-contact tilting in which the object slides to the right on the two supports and the contacts $A$ and $B$ are able to impart only the wrenches through the left edge of their respective friction cones. 

\subsubsection{$\{CB'_1B'_2\}$}
This configuration is depicted in Fig.~\ref{fig:mechanics}(b).
In the figure, the composite wrench cone of the three contacts is shown as the red and green shaded region labeled ``$+$'' or ``$-$'', formed by the left edge of the friction cones at $A$ and $B$ (denoted $\mathbf{f}^{A_1}$, $\mathbf{f}^{B_1}$) and the wrench applied at $C$ (denoted $\mathbf{f}_C$),
according to the method of moment labeling \cite{Mason__2001_3803}. In a quasistatic setting, the suggested manner of tilting is feasible because the wrench of gravity can be balanced: the wrench opposing it is contained in the composite wrench cone as the line of $-m\mathbf{g}$ has a positive (negative) moment with respect to the shaded region labeled ``$+$'' (``$-$'').



\subsubsection{$\{B'_1B'_2C\}$}
According to Fig.~\ref{fig:mechanics}(c), in this case the wrench of gravity cannot be balanced as the line of $-m\mathbf{g}$ does not yield a consistent moment with respect to the ``$-$'' labeled region.
The feasibility of tilting can be arranged by shrinking the ``$-$'' labeled region, by making contact $C$ ``sticky'' so that $\mathbf{f}_C$ can be directed upward.
Due to the large friction requirement, this manner of tilting can be considered less desirable than $\{B'_1CB'_2\}$ and $\{CB'_1B'_2\}$.

By comparing Figs.~\ref{fig:mechanics}(b) and \ref{fig:mechanics}(c) we can observe that
with a large (small) width-to-thickness ratio of the object model, the friction cone at $B$ sees larger (smaller) portion of
its edge containing $C$, resulting in an increased (decreased) availability to tilt in the manner of $\{B'_1CB'_2\}$ or $\{CB'_1B'_2\}$.
This suggests that objects that are relatively slender can be better suited to the task of tilting.
It can also be observed that
reorienting support \#2
such that $B$'s friction cone is redirected downward will facilitate three-contact tilting $\{CB'_1B'_2\}$.
This point was confirmed in our experiments (Sec.~\ref{subsec:experiments_arm}) as can be seen in the video attachment.

\subsection{Planning}
\label{subsec:planning}

Now we present a tilting strategy for keeping the object restrained
all the way through to a desired target configuration.

\begin{figure}[h]
    \centering
    \subfigure[]{
    \centering
    \includegraphics[trim={0cm 0cm 0cm 0cm}, clip, width=3.4in]{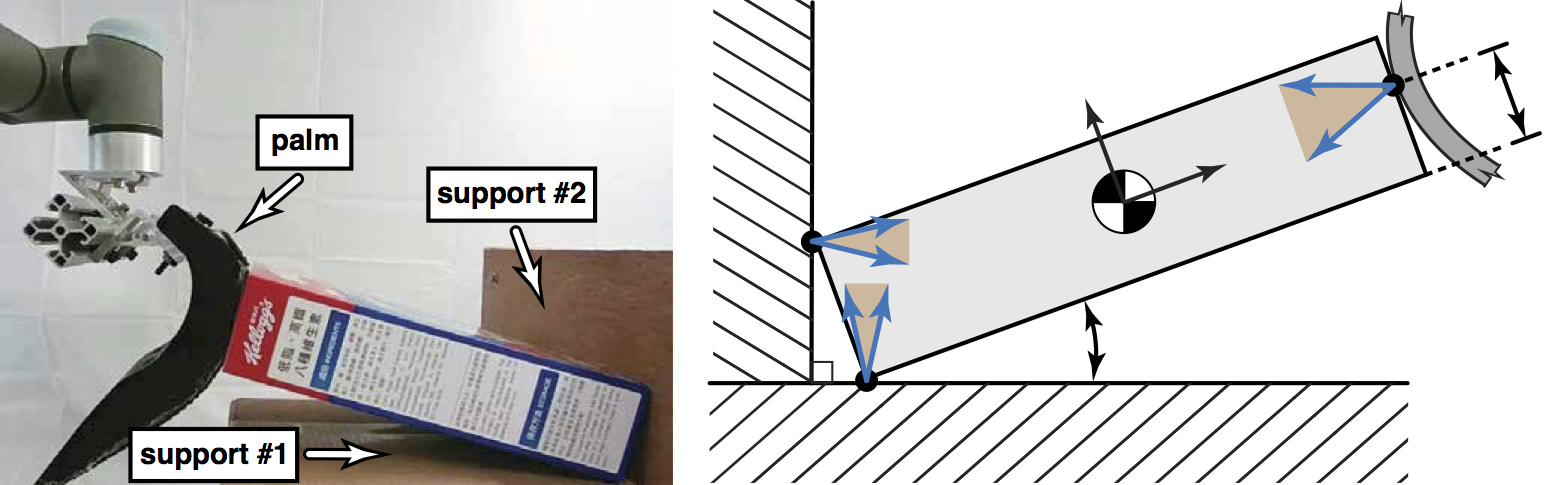}}
    \subfigure[]{
    \includegraphics[trim={0cm 0cm 0cm 0cm}, clip, width=3.3in]{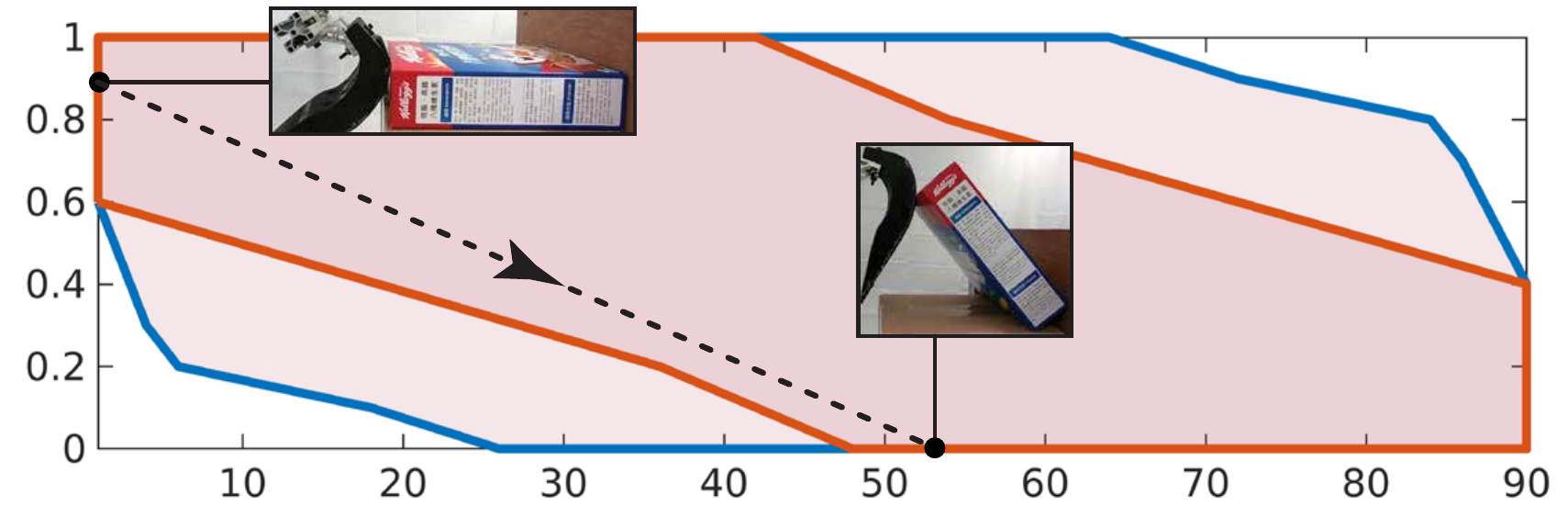}}
    \setlength{\unitlength}{1cm}
    \begin{picture}(0,0)(-5.75,-2.86)
    \put(-3.76,2.02){\footnotesize $\theta$}
    \put(-1.48,3.46){\footnotesize $\delta$}
    \put(-3.69,1.55){\footnotesize support \#1}
    \put(-5.89,4.02){\footnotesize support \#2}
    \put(-2.29,3.99){\footnotesize palm}
    \put(-5.33,1.65){\colorbox{white}{\makebox(0.2,0.07){\footnotesize $A$}}}
    \put(-5.82,2.63){\colorbox{white}{\makebox(0.16,0.07){\footnotesize $B$}}}
    \put(-2.36,3.19){\footnotesize $C$}
    \put(-3.23,2.86){\footnotesize $x$}
    \put(-3.78,3.34){\footnotesize $z$}
    \end{picture}
    \begin{picture}(0,0)(0.7,-6.95)
    \put(3.86,-6.59){\footnotesize $\theta\,(^\circ)$}
    \put(-3.65,-4.97){\footnotesize $\delta$}
    \put(2.55,-4.98){\rotatebox[origin=c]{-15}{\footnotesize $\mu_C=0.1$}}
    \put(2.78,-4.32){\rotatebox[origin=c]{-15}{\footnotesize $\mu_C=0.2$}}
    \put(-1.67,-4.53){\scriptsize initial configuration}
    \put(1.55,-5.6){\scriptsize target configuration}
    \end{picture}
    \caption{(a) Real tilting scenario (left) modeled in a plane normal to the two supports (right). Configuration variables $\theta$ and $\delta$ describe the progress of tilting.
    (b) The shaded area delimited by the red (blue) curve represents the set of configurations in which the object is in force-closure when friction coefficient at $C$ is $0.1$ ($0.2$). The plot is overlaid with a nominally feasible path for tilting. Friction coefficient at $A$ and $B$: $0.1$.
    }
    \label{fig:tilt_planning}
\end{figure}
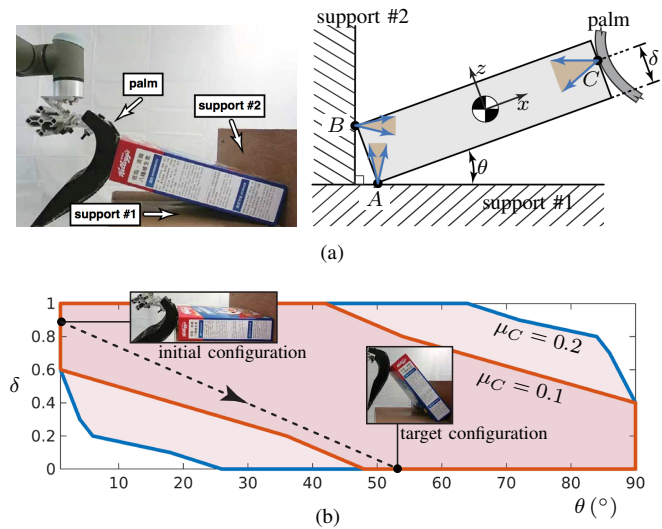

Recall
the picking operation shown in Fig.~\ref{fig:intro}.
A typical configuration of the object-environment system encountered during its
tilt phase (top row)
is shown in the left panel of Fig.~\ref{fig:tilt_planning}(a). The panel on the right schematically models it in a plane normal to the two supports.
According to this model, the configuration of the object-environment system is described by the angular displacement $\theta$ of the object, the location $\delta$ of object-palm contact $C$, and the angle between the two supports, which is set to be $90^\circ$.
It is expected that during tilting, the object rolls on the curved palm such that while $\theta$ is increasing, $\delta$ is decreasing monotonically.
Tilting supposedly begins (ends) when $\theta=0$ ($\delta=0$).

In order to find the set of configurations in which the object can be secured between the palm and the two supports, we test for force-closure using the method of linear programming \cite{lynch2017modern} for data points sampled uniformly in the $(\theta,\delta)$-space. Our software takes as input the object-environment configuration, the contact friction coefficients and the profile of the object model, and returns a binary result determining force-closure.
The shaded area delimited by the red (blue) curve in Fig.~\ref{fig:tilt_planning}(b) is computed by our software and corresponds to the set of configurations
in which the object model of Fig.~\ref{fig:tilt_planning}(a) is in force-closure
when the friction coefficient at $C$, denoted $\mu_C$, is assumed to be $0.1$ ($0.2$). The result indicates that increasing $\mu_C$ has the effect of enlarging the set of configurations in which the object can be in static equilibrium.
Fig.~\ref{fig:tilt_planning}(b) also suggests a way to plan the tilting maneuver: initial and target configuration of tilting can be chosen to lie on the $\delta$- and $\theta$-axis of the plot, respectively, such that the entirety of the straight-line path joining the two is contained in the shaded area.
This way of tilting will render the object force-closure-grasped during the manipulation process.
In the experiments to be presented, feasible initial and target configurations are chosen empirically to facilitate the picking operation (for example, to avoid collision between the palm and the bottom support).

\begin{figure}[h]
    \centering
    \subfigure[]{
    \includegraphics[trim={0 0 747 0}, clip, width=1.6in]{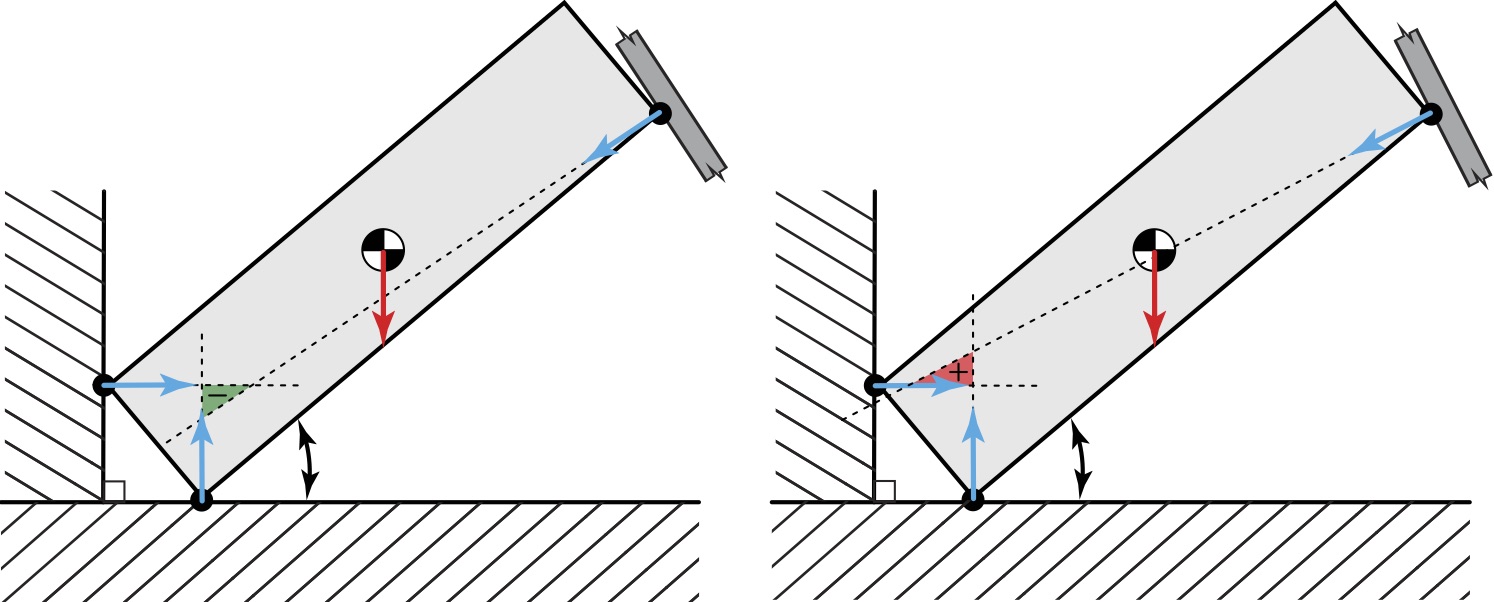}}
    \subfigure[]{
    \includegraphics[trim={747 0 0 0}, clip, width=1.6in]{media/mobility_target_config.jpg}}
    \setlength{\unitlength}{1cm}
    \begin{picture}(0,0)(-0.07,0)
    \put(-3.28,0.95){\colorbox{white}{\makebox(0.17,0.07){\footnotesize $A$}}}
    \put(-3.99,1.8){\colorbox{white}{\makebox(0.13,0.07){\footnotesize $B$}}}
    \put(-0.69,2.99){\footnotesize $C$}
    \put(-2.36,1.37){\footnotesize $\theta$}
    \put(-1.57,1.28){\footnotesize support \#1}
    \put(-4.12,2.95){\footnotesize support \#2}
    \put(-0.55,3.62){\footnotesize palm}
    \end{picture}
    \begin{picture}(0,0)(-4.39,0)
    \put(-3.28,0.95){\colorbox{white}{\makebox(0.17,0.07){\footnotesize $A$}}}
    \put(-3.99,1.8){\colorbox{white}{\makebox(0.13,0.07){\footnotesize $B$}}}
    \put(-0.69,2.99){\footnotesize $C$}
    \put(-2.36,1.37){\footnotesize $\theta'$}
    \put(-1.57,1.28){\footnotesize support \#1}
    \put(-4.12,2.95){\footnotesize support \#2}
    \put(-0.55,3.62){\footnotesize palm}
    \put(-5.25,0.25){\footnotesize $\theta=\theta'$, $\delta=0$}
    \end{picture}
    \caption{With an appropriately oriented palm at the target configuration, the object can be kinematically restrained from ungrasping by the way of rotating clockwise about a point in the green or red shaded region. First-order mobility analysis suggests that such ungrasping motion is feasible in (a) but not in (b).
    }
    \label{fig:mobility}
\end{figure}

At the target configuration ($\delta=0$), it is possible to kinematically restrain the object by taking into account the first-order geometry of the palm.
This is exemplified in Fig.~\ref{fig:mobility} by considering two identical target configurations, with the palm approximated as a straight segment.
In the figure, we examine how the orientation of the palm affects the mobility of the object.
According to Reuleaux’s method \cite{Mason__2001_3803}, the green (red) shaded region labeled ``$-$'' (``$+$'') formed by the contact normals at $A$, $B$ and $C$, represents the set of points about which the object can instantaneously rotate clockwise (counterclockwise).
This suggests that in the configuration shown in Fig.~\ref{fig:mobility}(a), it is possible for the object to {\em ungrasp} \cite{9681220} by rotating clockwise about a point in the ``$-$'' labeled region. In Fig.~\ref{fig:mobility}(b), such ungrasping is not kinematically feasible as the object is not allowed to penetrate into the palm. The critical orientation of the palm that separates the two outcomes can be determined by making the contact normals intersect at a common point.



\section{Picking By Tilting: Implementation, Experiments and Discussion}
\label{sec:experiments}

This section describes the implementation of our tilting-based picking technique and presents a set of experiments with a custom-made robotic device and a conventional robot arm.
See also the video attachment.

\subsection{Experimental Setup}

\begin{figure}[h]
    \centering
    \subfigure[]{
    \includegraphics[trim={0cm 0cm 0cm 0cm}, clip, height=1.16in]{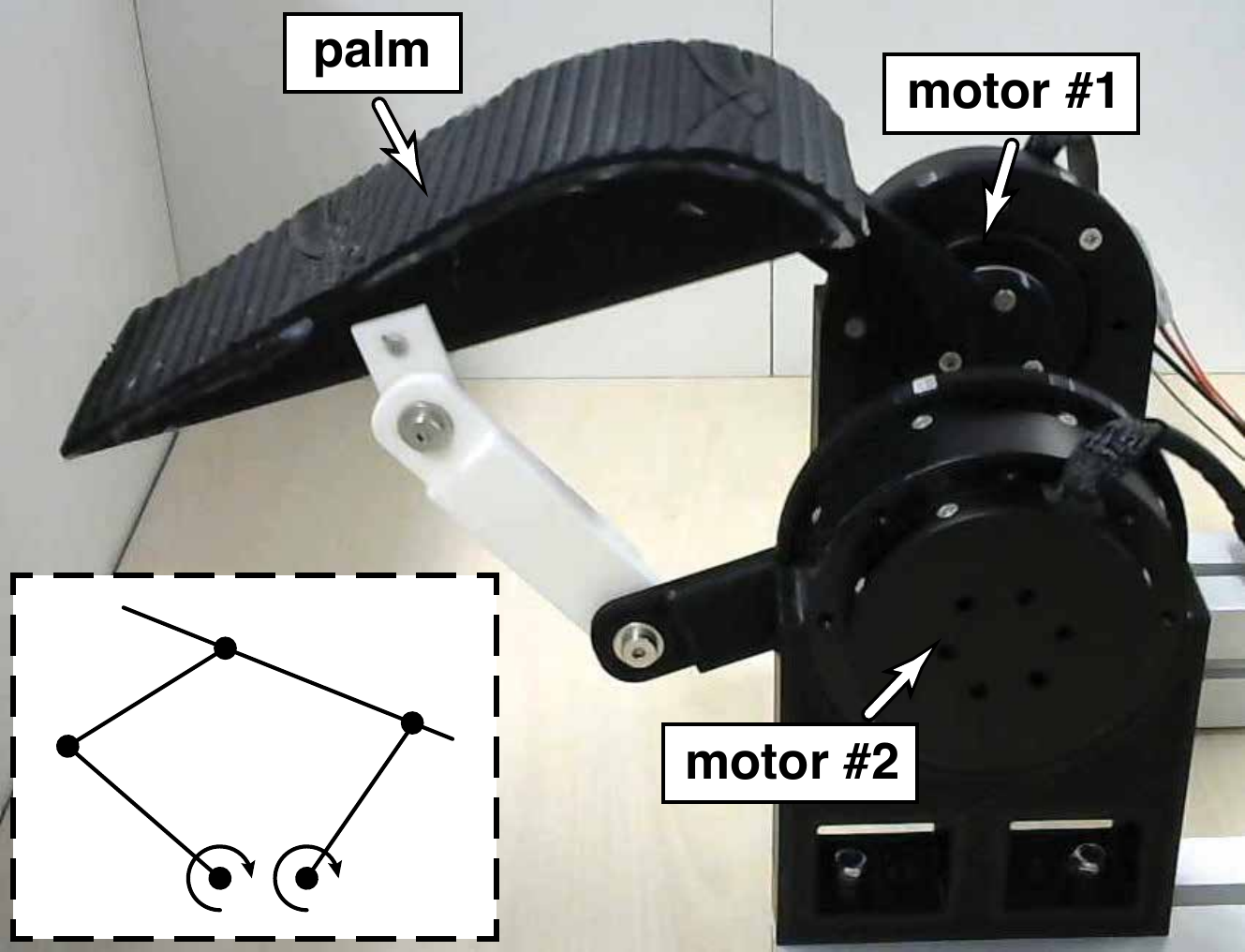}}
    \subfigure[]{
    \includegraphics[trim={0cm 0cm 0cm 0cm}, clip, height=1.16in]{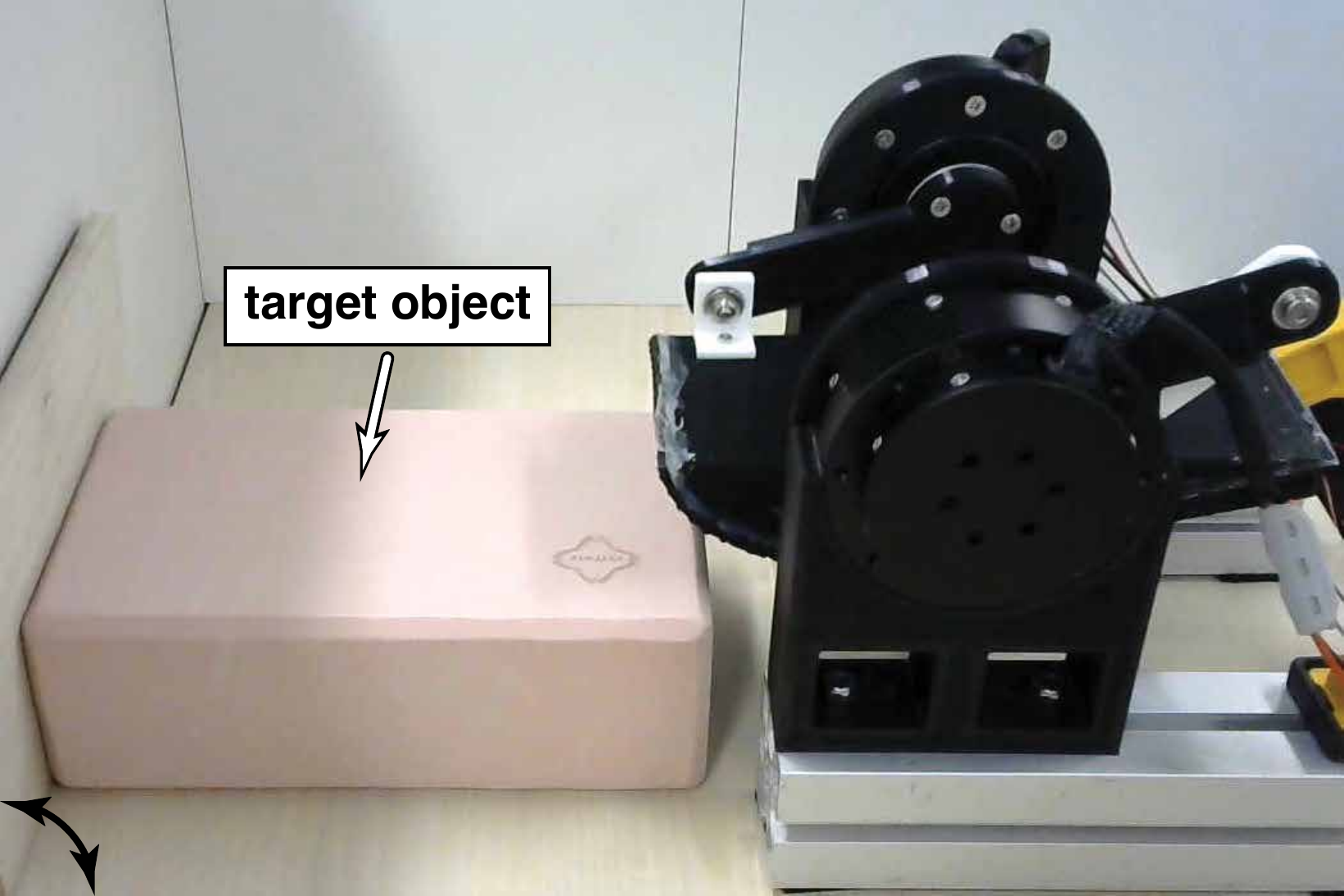}}
    \setlength{\unitlength}{1cm}
    \begin{picture}(0,0)(0,0)
    \put(0.26,0.72){\footnotesize $90^\circ$}
    \put(-3.33,0.72){\tiny motor \#1}
    \put(-4.07,0.72){\tiny motor \#2}
    \put(-3.69,1.68){\tiny palm}
    \end{picture}
    \subfigure[]{
    \includegraphics[trim={0cm 0cm 0cm 0cm}, clip, height=1in]{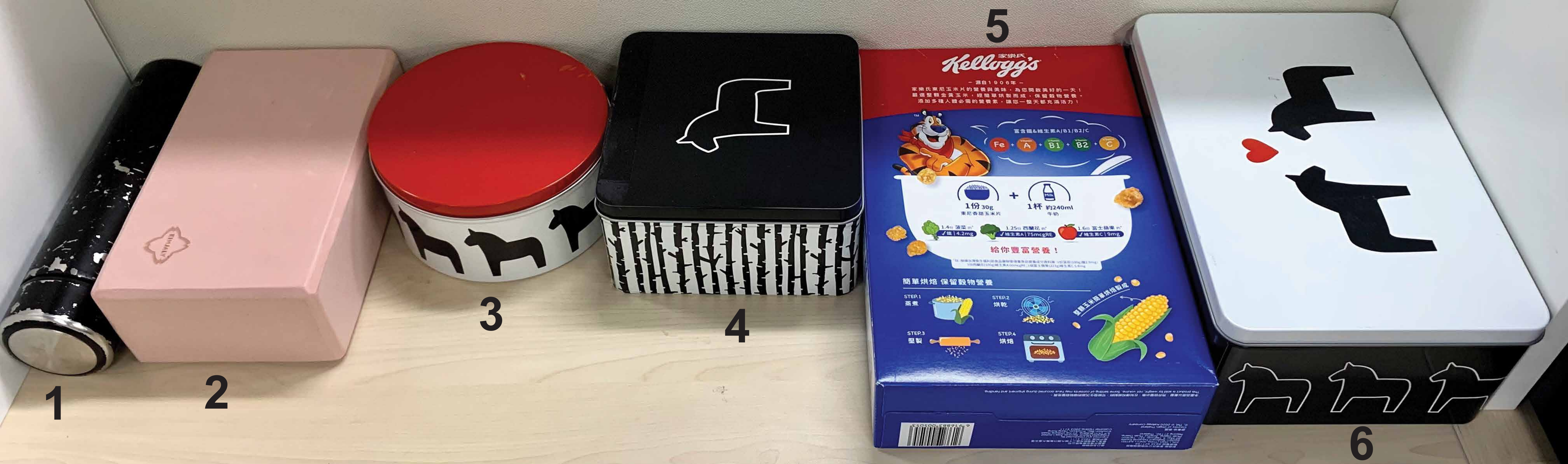}}
    \caption{(a) Our two-DOF robotic palm driven by a parallelogram linkage and two motors. (b) Hardware setup to pick an object in the corner of two perpendicular supports using our robotic palm. (c) Objects used in the experiments: 1-2 with two-DOF palm; 3-6 with UR3 arm (top-left in Fig.~\ref{fig:tilt_planning}).}
    \label{fig:2dof_palm_setup}
\end{figure}
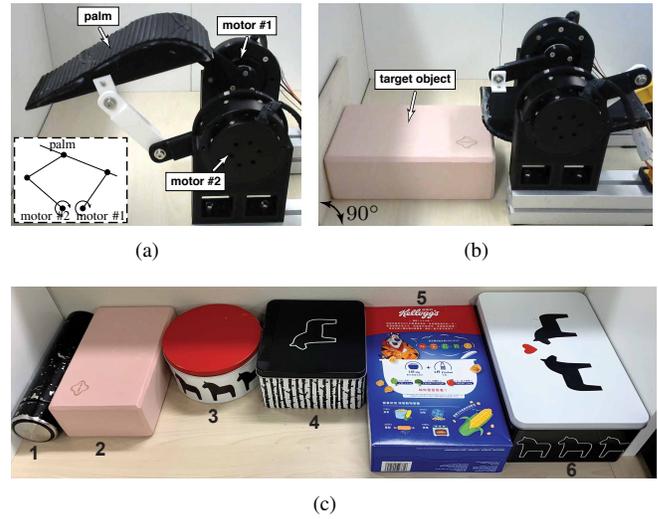

Fig.~\ref{fig:2dof_palm_setup}(a) shows our two-DOF robotic palm device driven by a closed-chain five-bar linkage mechanism \cite{10.7551/mitpress/2438.001.0001}.
The linkage is arranged in a parallelogram configuration such that its opposing link pairs are the same length.
Each of the two joints connected to the ground is actuated by a HT8108-J6 DC motor (rated torque: 6.9 Nm at 24 volts, according to the manufacturer\footnote{\url{http://www.haitaijd.cn}}) through a built-in planetary gear train.
The motors come with a driver that allows for position, speed and torque control.
The axes of the motors are horizontally offset to realize the desired manipulation behavior of the palm.
The palm is 3D printed and covered with a high-friction rubber material. Its proximal end features a curved, ellipsical surface to facilitate tilting by rolling. The distal end is either flat
or
customized to accommodate the object's bottom for lifting.
The overall experimental setup with our closed-chain robotic palm and a target object, a yoga block, in the corner of two perpendicular supports is shown in Fig.~\ref{fig:2dof_palm_setup}(b).
Another instance of our technique is demonstrated using a conventional six-DOF robot arm, Universal Robots UR3, by mounting a rigid body palm on its wrist (top-left in Fig.~\ref{fig:tilt_planning}).
No add-on force/torque sensor is used in our implementations.
Fig.~\ref{fig:2dof_palm_setup}(c) shows the different objects on which our method is tested.

Initially, the target object is placed in front of the
palm in a predefined pose, lying on support \#1 and touching support \#2. The manipulation for picking is supposed to happen in the plane normal to the two supports.
Constructively, pick two points $X$ and $Y$ on the curved end of the palm such that they are coincident with the object-palm contact $C$ at the beginning and end of the tilting, respectively,
assuming the object were to roll without slipping on $C$.
The course of picking as Fig.~\ref{fig:intro} is elaborated as follows.
First, the palm is gently slammed on the object to make contact at $X$.
Second, tilting is realized by jointly controlling the palm's motion and force: the palm is controlled to move upwards while
steadily pressing
the object against support \#2; its orientation is coordinated with its upward motion so that at the end of tilting, the object is kinematically restrained from ungrasping (recall Fig.~\ref{fig:mobility}). Third, the palm is rotated about $Y$ (which is now coincident with $C$) outwards, i.e., away from the robot base, so that its
distal
end is tucked into the gap between the object and support \#1. This step instantiates regrasping as the palm is reconfigured relative to the object.  Finally, the palm is lifted and retracted to detach the object from the supports.





\begin{figure*}[t]
    \centering
    \subfigure[]{
    \includegraphics[trim={0.5cm 24.3cm 0.5cm 0.5cm}, clip, width=6.6in]{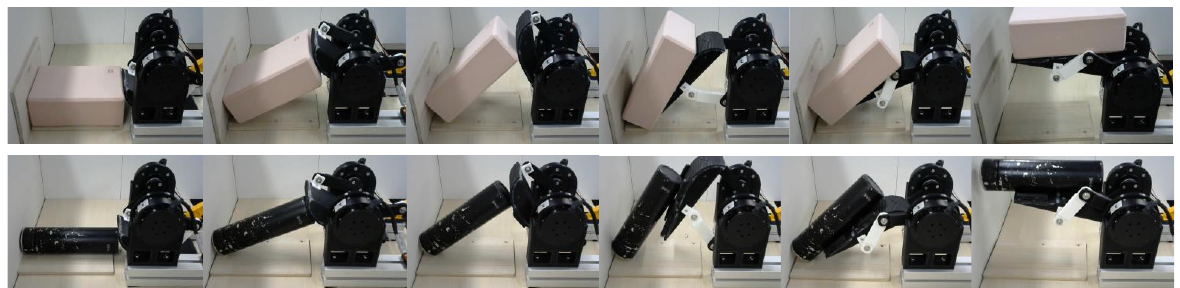}}
    \subfigure[]{
    \includegraphics[trim={0 0 0 0}, clip, width=6.4in]{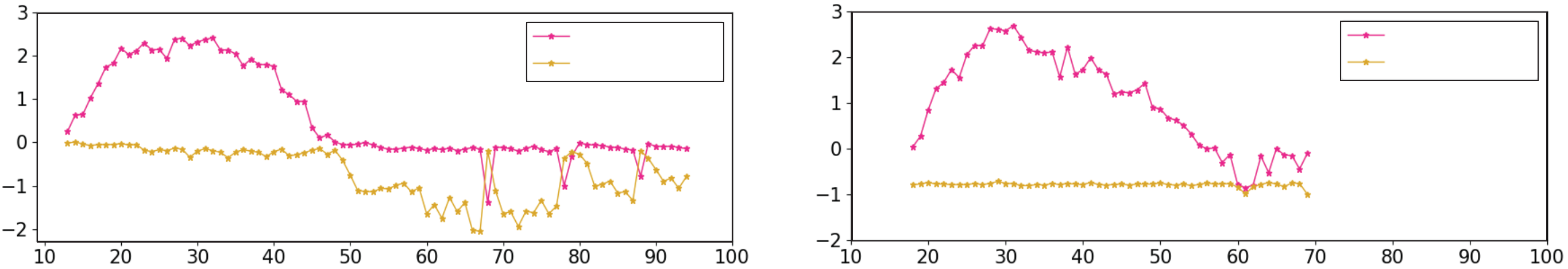}}
    \setlength{\unitlength}{1cm}
    \begin{picture}(0,0)(0,0)
    \put(-16.80,1.45){\footnotesize \rotatebox[origin=c]{90}{torque (Nm)}}
    \put(-8.33,1.45){\footnotesize \rotatebox[origin=c]{90}{torque (Nm)}}
    \put(-13.59,-0.27){\footnotesize command number}
    \put(-5.05,-0.27){\footnotesize command number}
    \put(-10.44,2.33){\footnotesize motor \#1}
    \put(-10.44,2.05){\footnotesize motor \#2}
    \put(-1.99,2.34){\footnotesize motor \#1}
    \put(-1.99,2.06){\footnotesize motor \#2}
    \put(-13.89,2.76){\footnotesize motion-controlled tilting}
    \put(-5.81,2.76){\footnotesize motion-force-controlled tilting}
    \end{picture}
    \subfigure[]{
    \centering
    \footnotesize
    \begin{tabular}{l M{2cm} c M{2.85cm} M{2.85cm} M{2.1cm}}
    \cmidrule{1-6}
    Object & Tilt control & Successes/Trials & Maximum torque (motor \#1, average, Nm) & Maximum torque (motor \#2, average, Nm) & Average number of commands\\
    \cmidrule{1-6}
    \multirow{2}{*}{\shortstack[c]{Yoga block (\#2, Fig.~\ref{fig:2dof_palm_setup}(c))}} & motion & 5/5 & 2.99 $\pm$ 0.08 & 1.87 $\pm$ 0.20 & 107 $\pm$ 10\\
    \cmidrule{2-6}
    &motion-force & 5/5 & 3.06 $\pm$ 0.03 & 0.83 $\pm$ 0.07 & 67 $\pm$ 3\\
    \cmidrule{1-6}
    \multirow{2}{*}{\shortstack[c]{Beverage container (\#1)}} & motion & 4/5 & 2.93 $\pm$ 0.04 & 1.51 $\pm$ 0.14 & 120 $\pm$ 8\\
    \cmidrule{2-6}
    &motion-force & 5/5 & 3.03 $\pm$ 0.05 & 1.00 $\pm$ 0.08 & 87 $\pm$ 2\\
    \cmidrule{1-6}
    \rule{0pt}{1mm} 
    \end{tabular}}
    \caption{Teleoperated object picking with our robotic palm. (a) Snapshots of experiment with a yoga block (top row) and a beverage container (bottom row). (b) Torque sensed by the motors during motion-controlled tilting (left) and motion-force-controlled tilting (right). (c) Summary of results.
    }
    \label{fig:5bar_picking_up_process}
\end{figure*}

\subsection{Experiments with Our 2-DOF Robotic Palm}

We first tested the picking technique with our custom-made robotic palm. Two types of implementation were considered to perform the tilting maneuver: motion-controlled, in which the angular position of each motor was commanded, and motion-force-controlled, in which the angular position (torque) of motor \#1 (motor \#2) was commanded. In both settings, the motors were controlled in a teleoperated manner by the human experimenter.

The successive snapshots in Fig.~\ref{fig:5bar_picking_up_process}(a) show successful picking experiments with two objects (yoga block \#2 and beverage container \#1), when the tilting maneuver was motion-force-controlled.
During the tilt phase (between the first and third panel), the angular position of motor \#1 was commanded to increase monotonically, while the torque applied by motor \#2 was kept constant.
This enabled the palm to tilt up the object while steadily pressing it against support \#2 and at the same time, reorient itself.
The end of tilt phase was determined manually by the experimenter (it can also be detected autonomously using the proprioceptive information contained in the motor encoders) and the controller was switched off.
Subsequently, the palm was motion-controlled to regrasp and lift up the object (panels fourth to sixth).

Fig.~\ref{fig:5bar_picking_up_process}(b) compares the torque sensed by the motors during the tilt phase of the picking experiment with object \#2, when the tilting maneuver was motion-controlled (left) and motion-force-controlled (right).
The plots in the figure indicate that compared to the pure motion control, which required careful maneuvering to maintain contact with the object, tilting could be arranged with fewer teloperation commands with the simpler motion-force control.
In addition, pure motion control rendered it difficult to steadily press the object and resulted in larger fluctuations in the torque sensed by the motors.
Here, the maximal torque recorded was also much larger than that recorded with the motion-force control
(2.05Nm vs. 0.99Nm on motor \#2; 2.41Nm vs. 2.69Nm on motor \#1),
suggesting that the hybrid control setting outperformed in terms of maintaining a ``gentle'' grasp of the object.

See the table in Fig.~\ref{fig:5bar_picking_up_process}(c) summarizing the results of our experiments.



\begin{table}[h]
\small
\centering
\caption{Picking experiments performed with UR3 arm}
\label{tab:arm_experiments}
\begin{tabular}{l M{2cm} M{2.cm}}
\cmidrule{1-3}
Object & Angle between supports ($^\circ$) & Successes/Trials \\
\cmidrule{1-3}
Steel box (\#4, Fig.~\ref{fig:2dof_palm_setup}(c)) & $90^\circ$ & 0/5 \\
Steel box (\#4) & $61^\circ$ & 5/5 \\
Steel box (\#6) & $61^\circ$ & 3/5 \\
Steel box (\#3) & $61^\circ$ & 5/5 \\
Cardboard box (\#5) & $90^\circ$ & 5/5 \\
\cmidrule{1-3}
\end{tabular}
\vspace{-0.4cm}
\end{table}

\begin{figure*}[t]
    \centering
    \subfigure[]{
    \includegraphics[trim={0 0 0 0}, clip, width=6.6in]{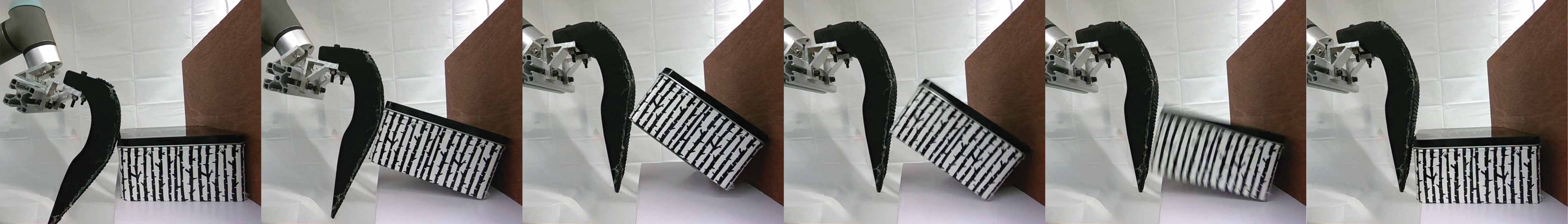}}
    \subfigure[]{
    \includegraphics[trim={0 0 0 0}, clip, width=6.6in]{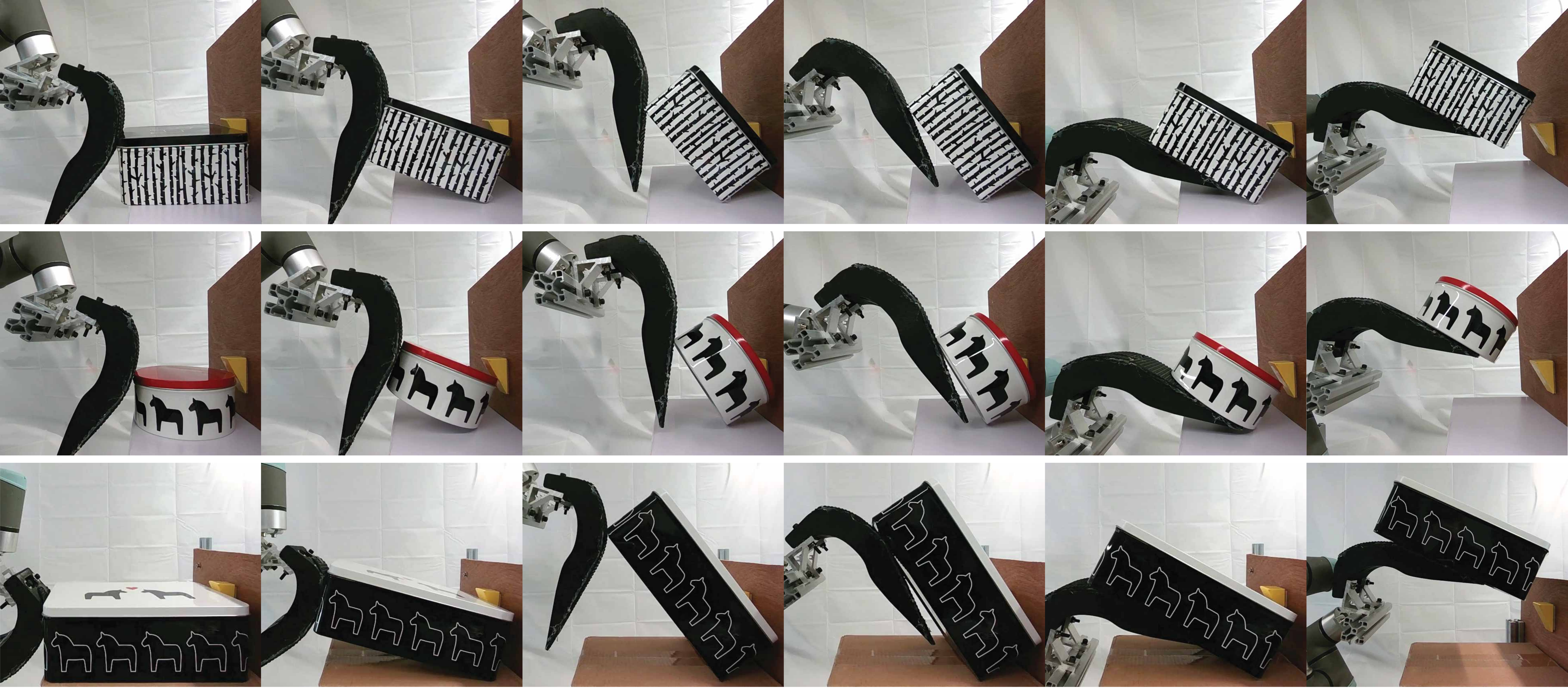}}
    \caption{Object picking with a conventional manipulator. Snapshots of (a) unsuccessful, and (b) successful picking experiments.}
    \label{fig:arm_exps}
\end{figure*}

\subsection{Experiments with a Conventional Manipulator}
\label{subsec:experiments_arm}

We also tested our picking technique using a conventional manipulator arm UR3, equipped with a rigid body palm to interact with the object (top-left in Fig.~\ref{fig:tilt_planning}). Here, the robot executed a programmed trajectory to perform the picking operation.
But the high reflected inertia of the robot arm rendered it difficult to
steadily press
the object during tilting.
Fig.~\ref{fig:arm_exps}(a) shows a common failure mode witnessed during the experiments with a steel box object (first row in Table \ref{tab:arm_experiments}).
Towards the end of tilting (third panel), the object is wedged between the palm and support \#2.
Subsequently, however, the frictional contact wrenches cannot balance the wrench of gravity and the object falls on the bottom support (last panel).
This outcome was also observed when the experiment was repeated with objects \#3 and \#6 of Fig.~\ref{fig:2dof_palm_setup}(c) (see \cite{8794366} reporting similar failure mode).

One way to resolve this failure mode is to reorient support \#2 such that the contact normal for its contact with the object is redirected downward. 
This will physically suppress the object from breaking contact with support \#1 during tilting (recall the discussion in Sec.~\ref{subsec:mechanics}).
In our experiments, this is achieved by using the inclined face of a right triangular prism as support \#2. This fixture design solution, in which the angle formed by the two supports is made more acute, resulted in more successful object picking (see Fig.~\ref{fig:arm_exps}(b), rows 2-4 in Table \ref{tab:arm_experiments}).

Lastly, we note that the failure mode described above was not witnessed when experimenting with the more deformable cereal box object made of cardboard material (Fig.~\ref{fig:intro}, last row in Table \ref{tab:arm_experiments}). It is hypothesized that in this case, compliance in contacts and local contact curvature surrounding the deformed shape of the object might have enabled more secure tilting.

\subsection{Discussion}

In our experiments, the planning blueprint presented in Sec.~\ref{subsec:planning} is realized in an open-loop manner, considering the lack of sensing capability to keep track of the object-environment configuration.
Thus, when the tilting maneuver is performed,
the actual path traced by the object-environment system in the $(\theta,\delta)$-space
differs from the nominally feasible path planned as in Fig.~\ref{fig:tilt_planning}(b).
It is possible to close the gap by coordinating the passive and active interactions happening at the object-palm contact interface (the location of object-palm contact evolves as the object passively ``complies'' to the shape of the palm while it is being tilted, as well as through the active reorientation of the palm carried out in order to kinematically restrain the object by the end of tilting).
Modeling these interactions will necessitate taking into consideration the geometry of the palm, which is ignored in the current work.
Nevertheless, as predicted in Sec.~\ref{subsec:planning}, the use of high-friction material on the palm seems sufficient to attain the high success rates reported in the experiments.

\section{Conclusion}
In this paper, we presented a robotic in-hand manipulation technique that can be applied to pick objects that are too large to grasp in a prehensile manner, by taking advantage of their interaction with the environmental surfaces.
We have demonstrated the effectiveness of our technique through a set of experiments performed with a custom-made robotic device and conventional manipulator arm. Our experiment results showed that object picking with our method can generally be performed reliably using simple hardware and control.

Possible directions for future work include: (1) generalization of the task mechanics to incorporate the geometry of the end-effector; (2) autonomous execution of the manipulation technique from object recognition to successful object picking; (3) robust execution of the manipulation plan by incorporating tactile feedback. 



\section{Acknowledgments}
This work is supported by InnoHK of the Government of Hong Kong via the Hong Kong Centre for Logistics Robotics and is also supported by the CUHK T Sone Robotics Institute. The authors thank Linzhu Yue and Zhitao Song for the motors and driver software provided by them.

\bibliographystyle{IEEEtran} 
\bibliography{IEEEabrv,IEEEexample}

\end{document}